# A Novel Gripper with Semi-Peaucellier Linkage and Idle-Stroke Mechanism for Linear Pinching and Self-Adaptive Grasping *

Haokai Ding and Wenzeng Zhang, *Member, IEEE*

*Abstract*—This paper introduces a novel robotic gripper, named as the SPD gripper. It features a palm and two mechanically identical and symmetrically arranged fingers, which can be driven independently or by a single motor. The fingertips of the fingers follow a linear motion trajectory, facilitating the grasping of objects of various sizes on a tabletop without the need to adjust the overall height of the gripper. Traditional industrial grippers with parallel gripping capabilities often exhibit an arcuate motion at the fingertips, requiring the entire robotic arm to adjust its height to avoid collisions with the tabletop. The SPD gripper, with its linear parallel gripping mechanism, effectively addresses this issue. Furthermore, the SPD gripper possesses adaptive capabilities, accommodating objects of different shapes and sizes. This paper presents the design philosophy, fundamental composition principles, and optimization analysis theory of the SPD gripper. Based on the design theory, a robotic gripper prototype was developed and tested. The experimental results demonstrate that the robotic gripper successfully achieves linear parallel gripping functionality and exhibits good adaptability. In the context of the ongoing development of embodied intelligence technologies, this robotic gripper can assist various robots in achieving effective grasping, laying a solid foundation for collecting data to enhance deep learning training.

## I. INTRODUCTION

Grasping, a fundamental human skill that encompasses parallel clamping and adaptive enveloping, serves as a driving force in the research of anthropomorphic robotic grippers. Contemporary research analysis systematically classifies the evolution of anthropomorphic robotic manipulators into three primary categories: Simplified industrial end-effectors, which feature streamlined structural configurations and enhanced payload capacities within the 50-200 N range, albeit with operational limitations confined to precision pinch-grasp modes; Biomimetic dexterous manipulators, which emphasize morphological fidelity to the human hand through multi-digit articulation systems with ⩾3 fingers per unit and ⩾3 degrees of freedom (DOF) per phalangeal joint, with notable examples including the Stanford/JPL Hand [1], Utah/MIT Hand [2], DLR/HIT Hand [3], Robonaut Hand [4], and CBPCA Architecture [5]; and Underactuated adaptive grippers, which address the inherent limitations of fully-actuated systems through synergistic mechanical optimizations, achieving 30-50% mass reduction through structural simplification and 60-80% cost-efficiency improvements compared to their traditional dexterous counterparts [6].

Despite demonstrating advanced kinematic flexibility and high-fidelity manipulation capabilities, dexterous robotic hands continue to encounter significant implementation barriers in real-world deployment scenarios. The electromechanical configuration, which necessitates dedicated actuators for each degree of freedom (DOF), introduces prohibitive levels of mechanical complexity and volumetric constraints. Additionally, the substantial manufacturing expenditures associated with this configuration make achieving human-level dexterity technologically challenging. These systemic challenges have spurred the emergence of underactuated architectures as promising alternatives, garnering considerable research attention due to their inherent mechanical compliance and reduced actuation requirements.

Underactuation mechanisms and their applications in robotic grasping have emerged as a critical research frontier in the field of robotics. Fueled by the escalating demands for industrial automation, operation in unstructured environments, and reliable task execution, underactuated systems exhibit significant potential due to their structural simplicity, cost-effectiveness, and enhanced environmental adaptability. Key advancements in this area include: the MARS hand [7] developed by Gosselin, which integrates parallel clamping and adaptive grasping dual modalities; the SARAH hand [8], the first underactuated manipulator to operate aboard the International Space Station; the SDM hand [9], a reconfigured four-digit underactuated system from Dollar; the Omega (Ω) Gripper [10], which combines passive compliance with under-actuation for omnidirectional object manipulation; Zhang pioneering series [11], [12], featuring indirect/coupled adaptive transmission mechanisms; economical four-finger underactuated gripper of Anhui University [13], which utilizes serial differential kinematics for precision fingertip control; tri-modal gripper by Chen [14], employing Chebyshev linkage configurations; systematic classification framework by Feng [15] for underactuated digit topologies; Tesla tendon-actuated hand [16], an industrial-grade device that coordinates 11 joints through six motors with tendon-sheath transmission and torsional spring reset mechanisms; and Tencent innovative underactuated platform [17], which achieves a 92% success rate in complex scenarios through dynamic 6-DOF motion control strategies.

This paper introduces a novel SP mechanism that is based on an optimized design of the Peaucellier mechanism. Building upon this foundation, an innovative underactuated delayed adaptive robotic gripper, named the Semi-Peaucellier linkage and Delayed transmission (SPD) gripper (shown in Fig. 1), is developed. The design achieves linear parallel gripping by integrating an improved Peaucellier linkage

* Research supported by the *Foundation of Enhanced Student Research Training (E-SRT)* and *Open Research for Innovation Challenges (ORIC)*, X-Institute.

Haokai Ding is with Future Technology School, Shenzhen Technology University and Laboratory of Robotics, X-Institute, Shenzhen, China.

Wenzeng Zhang is with Laboratory of Robotics, X-Institute, Shenzhen, China and Dept. of Mechanical Engineering, Tsinghua University, Beijing, China. (Corresponding author, email: zhangwenzeng@x-institute.edu.cn).

mechanism with a parallelogram mechanism. Furthermore, it enables hybrid switching between linear parallel gripping and adaptive gripping modes through idle stroke transmission. Section II provides a detailed explanation of the grasping mode implementation. Section III describes the design of the SPD finger mechanism. Section IV presents a mechanical error analysis. Section V reports on the grasping experiments conducted. Finally, conclusions are summarized in Section VI.

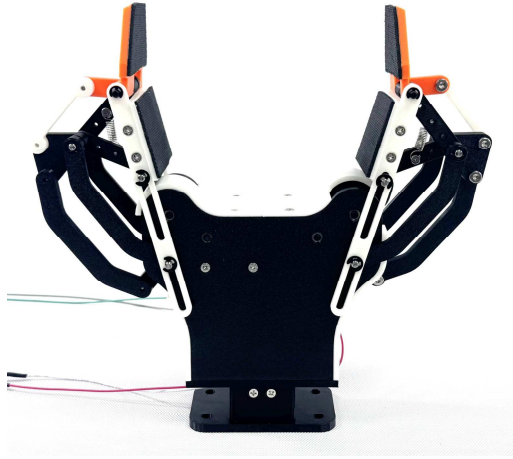

Figure 1. The SPD gripper.

## II. DESIGN OF THE SP MECHANISM

### A. Linear Fingertip Motion: Semi-Peaucellier Linkage

The operational principle of the classical Peaucellier mechanism[18], illustrated in Fig. 2a, demonstrates that the conventional eight-bar linkage configuration generates precise linear motion at node D through geometric constraint propagation. Kinematic analysis confirms that this mechanical system achieves straight-line trajectory formation via its characteristic linkage arrangement.

The triangles $\Delta EC_1C_2$, $\Delta BC_1C_2$, and $\Delta DC_1C_2$ are all isosceles with $C_1C_2$ as their common base, where the respective vertices E, B, and D serve as apex points. Consequently, the vertices E, B, and D are necessarily located on the perpendicular bisector of $C_1C_2$, establishing their collinearity due to the shared geometric constraint. This bisector simultaneously acts as the altitude, angle bisector, and median from the apex to the base in each isosceles triangle configuration. For the triangles $\Delta EFC_1$ and $\Delta DFC_1$, analogous geometric relationships may be derived based on the properties of their respective vertex angles and side congruencies.

$$EC_1^2 = FE^2 + FC_1^2 \quad (1)$$

$$DC_1^2 = DF^2 + FC_1^2 \quad (2)$$

$$EC_1^2 - DC_1^2 = FE^2 - DF^2 = DE \times BE \quad (3)$$

Given that both $EC_1$ and $DC_1$ remain constant, the product $DE \times BE$ is consequently invariant, which satisfies the necessary condition for achieving precise straight-line motion. Consequently, point D consistently traces a linear trajectory perpendicular to AE.

The classical Peaucellier linkage is constrained by inherent limitations in practical implementations: the mandatory geometric constraints of collinear alignment (specifically, the D-B-E node alignment in the trajectory of point E) generate kinematic conflicts that substantially degrade clamping linearity. To address this limitation, a Semi-Peaucellier linkage (SP) is developed through systematic topological reconstruction, as illustrated in Fig. 2b.

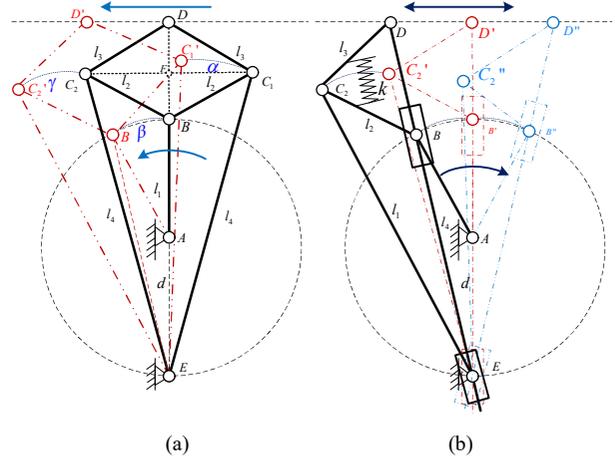

(a)                    (b)

Figure 2. Peaucellier linkage and the Semi-Peaucellier mechanism.

This innovative design integrates the collinear nodes D-B-E into a unified ternary element and implements sliding joints at positions B and E. Experimental observations revealed that gravitational forces induce vertical displacement in the sliding joints, thereby compromising linear motion accuracy. To mitigate this effect, a tension spring is introduced between links $l_2$ and $l_3$ to stabilize the relative positions of the sliders. Consequently, the proposed SP reduces the traditional seven-bar architecture to a five-component configuration, achieving structural simplification while maintaining motion precision.

The proposed topological reconstruction achieves a 37.5% reduction in component count while effectively mitigating mechanical interference constraints, thereby enabling the practical realization of linear parallel clamping mechanisms that conventional configurations could not implement.

### B. Orientation-Stable Fingertip Control: Parallelogram Linkage Synthesis for Parallel Grasping

A Semi-Peaucellier Underactuated Grasping finger mechanism was proposed to ensure the gripper segment maintains a consistent orientation relative to the base during linear motion to facilitate parallel grasping. The fingertip trajectory remains aligned along a straight line throughout the entire movement, as illustrated in Fig. 3. This design ensures that the gripper maintains a fixed orientation relative to the base, which is critical for achieving stable and precise grasping operations. The double parallelogram mechanism consists of two identical parallelograms connected in series, which effectively cancel out any rotational motion while allowing pure translational displacement.

The optimized structural configuration reduces complexity in control system implementation while improving operational reliability through significant reduction of angular displacement in grasping operations. Additionally, the symmetrical kinematic design promotes uniform load distribution throughout the gripper components, thereby mitigating potential alignment deviations during dynamic manipulation tasks. Experimental validation demonstrates superior capability of dual-parallelogram Mechanism in preserving planar orientation integrity, rendering it advantageous for precision-critical implementations including microscopic assembly systems and fragile material handling apparatuses.

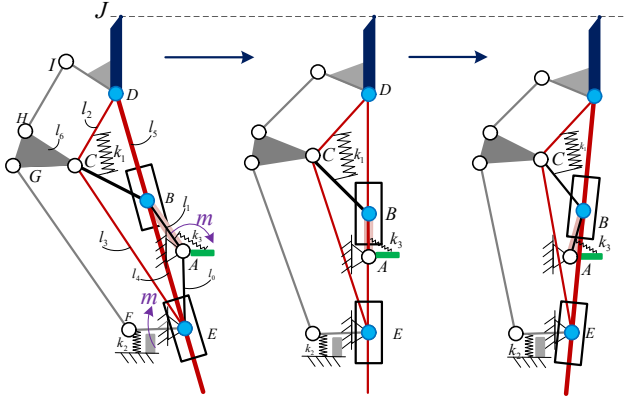

Figure 3.  Structure and Linear Motion of the SPD finger.

Additionally, the compact and lightweight design of the mechanism enables its integration into various robotic systems without significantly increasing the overall payload or complexity. The theoretical analysis and practical implementation of this mechanism are further discussed in the subsequent sections, providing a comprehensive under-standing of its design principles and operational advantages.

## III. Design of The SPD gripper

The SPD gripper comprises two symmetrically configured grasping digits, with their actuation principles to be systematically examined in subsequent chapters. This symmetrical kinematic architecture ensures balanced load distribution and repeatable operational characteristics during object manipulation phases, a fundamental requirement for maintaining system stability and millimeter-level positioning precision. Furthermore, the kinematically synchronized digit configuration enables morphological compliance with polymorphic workpiece profiles while preserving high-fidelity contact retention throughout grasping cycles. The forthcoming analytical framework will delineate the governing principles of both trajectory planning algorithms and contact dynamics that define the operational envelope of the articulated mechanism.

### A. Structural Configuration

The finger mechanism integrates two core components: a Semi-Peaucellier linkage and a double parallelogram mechanism (DPM), as illustrated in Fig. 4.

The mechanical system consists of three principal elements: the proximal phalange (1st P), distal phalange (2nd P), and cam-driven coupling mechanism (CDL). Through a distal joint shaft, 1st P and 2nd P are kinematically connected, permitting 2nd P to achieve planar translation relative to the fixed base while preserving autonomous rotational capability.

During static operation, torsional spring $S_2$ mechanically couples intermediate linkages $ML_2$ and $ML_4$ to maintain spatial congruence between proximal and medial joint axes. This design methodology significantly reduces gravitational perturbation effects on linear guideways Slat 1-Slat 2, consequently enabling stable rectilinear motion generation in the SP actuator assembly. Actuation of the SPD module is accomplished through a cable-driven transmission system employing driving linkage (DL).

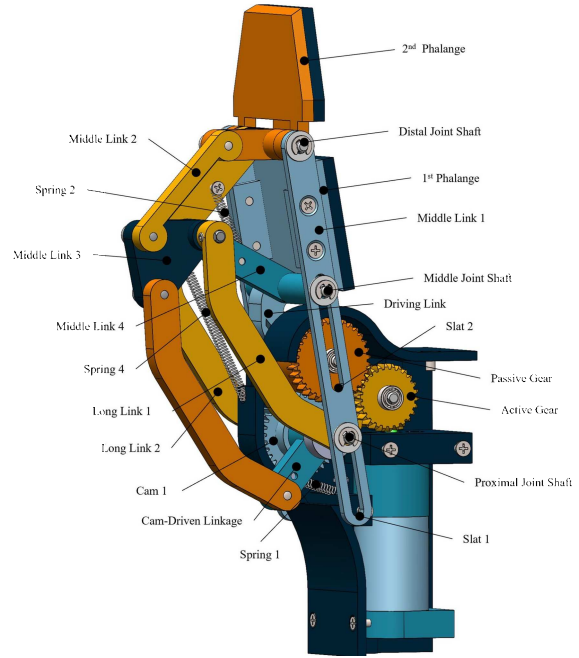

Figure 4.  The Design of the SPD finger.

The coordinated interaction between central joint axis of $ML_1$ and proximal joint axis ensures precise linear motion transfer, with both components maintaining constrained sliding motion within precision-machined guide channels. DL operation generates programmable linear displacement in 2nd P via force transmission through medial linkage $ML_4$.

Compression spring S1 establishes a mechanical pathway between the base mounting point and the CDL, effectively compensating for kinematic clearance during bidirectional linear actuation sequences while ensuring consistent motion uniformity. Tension spring 4 is connected between the two sides of $ML_3$ and the base to eliminate the non-uniform motion of the SPD mechanism during the opening action.

### B. Idle-stroke mechanism

The differential parallel mechanism (DPM) integrates an optimized four-bar linkage system featuring extended structural members long link 1/ long link 2 paired with intermediate connectors $ML_2$/$ML_3$, where $ML_3$ adopts a triangular prism geometry. This constrained parallel kinematic configuration provides millimeter-level motion tracking accuracy through coordinated linkage movements.

The SPD mechanism achieves adaptive enveloping of objects through active control, as illustrated in Fig. 5b.

Computational analysis reveals that during initial motion, the angular displacement between Cam 1 (C1) and the toggle block of CDL measures 90°.

During standard parallel grasping operations, the rotational movement of $C_1$ progressively reduces this angular interval until complete digital closure occurs, whereupon the dual toggle blocks establish mechanical contact. This kinematic interaction drives $C_1$ to actuate CDL, inducing deformation in the parallelogram linkage. Consequently, Middle Link 4 ($ML_4$) undergoes compressive loading, which transfers operational force to Spring 3. This force transmission mechanism ultimately induces pronation of the entire finger assembly, thereby realizing active adaptability.

The SPD finger mechanism incorporates dual transmission systems. The primary transmission system - comprising a worm gear set, active gear, passive gear, and actuation rod - implements parallel grasping functionality. The motor transmits torque through the worm gear set to the active gear, which subsequently engages the passive gear to drive the actuation rod. The arc motion of the actuation rod is converted into linear motion of the secondary phalange.

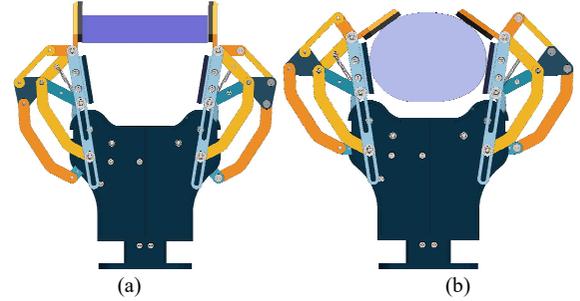

Figure 6. Two grasping modes of the SPD gripper. (a) Linear parallel pinching for precision grasp; (b) Self-adaptive grasping for power grasp.

The secondary transmission system - consisting of a worm gear set, driving gear, driven gear, $C_1$, and CDL mechanism - enables active-adaptive grasping. Motor torque is transferred through the worm gear set to sequentially activate the driving gear, driven gear, and $C_1$. A 90° idle stroke exists between $C_1$ and the CDL mechanism. During parallel grasping, the CDL mechanism rotates without cam block engagement. When finger mechanisms converge or encounter objects, cam wheel 1 rotation triggers cam block contact, propelling the CDL mechanism. This deformation in the parallelogram mechanism (where the secondary phalange connects to the upper side of the dual-parallelogram kinematic chain) induces active inward bending of the secondary phalange through dual-parallelogram transmission, achieving shape adaptation.

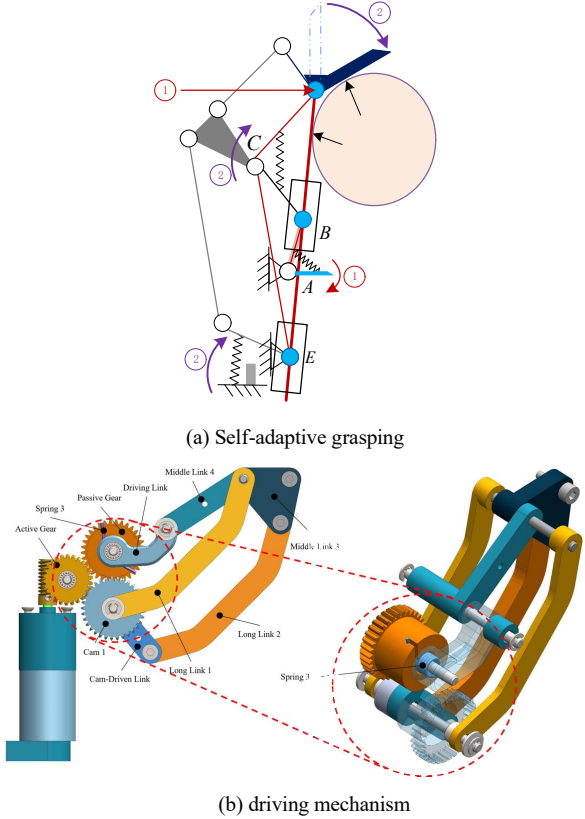

(a) Self-adaptive grasping

(b) driving mechanism

Figure 5. Self-adaptation and the driving mechanism, the Idle-stroke mechanism between Cam 1 and Cam-Drivern Link.

## C. Operational Modes

The SPD gripper demonstrates dual-mode grasping capability, as illustrated in Fig. 6: a) The parallel grasping mode (Fig. 6a) is activated when manipulating relatively regular elongated objects and small standardized-shaped items, maintaining parallel fingertip alignment through constrained kinematic chains; b) The active-adaptive grasping mode (Fig. 6b) operates during handling of large irregular objects, exhibiting passive compliance through active self-adjusting rotation of secondary phalanges to achieve optimal shape conformity during experimental evaluations. These two operational modes ensure stable object interaction through distinct mechanical constraints and compliance mechanisms.

The finger mechanism demonstrates two distinct operational modes, as shown in Fig. 7. The underlying kinematic analysis reveals:

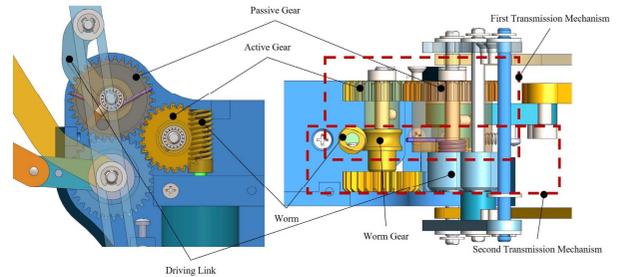

Figure 7. Driving mechanism of the SPD finger, divide the driving power into two lines and take separate routes.

## IV. MAIN ERROR ANALYSIS OF THE SPD FINGER

Based on the kinematic characteristics of the improved Peaucellier mechanism, an error analysis framework is established as illustrated in Fig. 8. The primary sources of error can be categorized as follows: geometric manufacturing errors ($\Delta L$), where machining tolerances of linkages cause distortion in proportions of the mechanism; joint clearance (c), where the clearance in slider-guide interfaces induces random trajectory vibrations; frictional nonlinearity ($\mu$), where Coulomb friction at sliding contact surfaces leads to velocity-dependent hysteresis; and assembly errors ($\Delta x$, $\Delta y$), where misalignments in hinge point positioning result in systematic error accumulation; the angle between the 1st linkage and the slider ($\theta$).

The actual displacement in the X direction, considering manufacturing errors, is calculated by:

$$x_{real} = L_{real} \cdot (1 - \cos(\theta)) \quad (4)$$

where $L_{real} = L + \Delta L$ is the actual linkage length, and $\Delta L$ is the manufacturing error.

The total error in the Y direction is the sum of geometric error, friction error, clearance error, and random error. Each error component is described below. The geometric error is calculated by:

$$geo\_error = 0.3 \cdot \sin(2\theta) + \frac{L_{real} - L}{L} \cdot ideal\_y \quad (5)$$

where the first term represents the angle error, and the second term represents the linkage length error.

The friction error is calculated by:

$$friction\_error = 0.1 \cdot \theta \quad (6)$$

The clearance error is calculated by:

$$clearance\_error = 0.05 \cdot clearance \cdot \theta^2 \quad (7)$$

The random error is calculated by:

$$random\_error = 0.2 \cdot N(0,1) \quad (8)$$

Based on the iterative computational process outlined in Equations (4)-(8), Fig. 7 demonstrates the motion trajectory and error characteristics of the SPD finger.

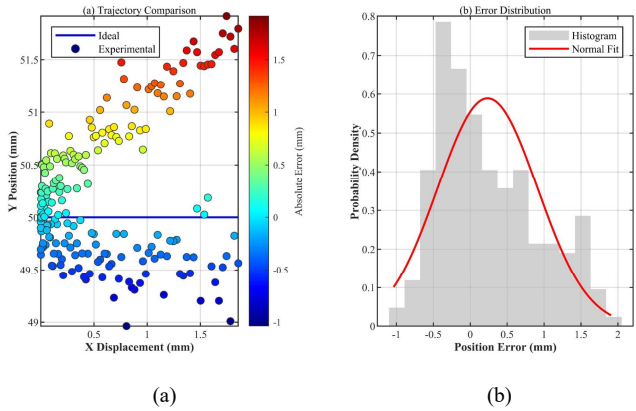

(a)                  (b)

Figure 8. Analysis of trajectory and error: (a) Comparison of theoretical and experimental trajectories, (b) Probability distribution of vertical errors.

Table I quantifies the experimental measurements and theoretical influence coefficients of various error parameters. Among them, geometric error exhibits the highest sensitivity (S=0.42), confirming sensitivity of the mechanism to the precision of linkage dimensions which sensitivity values were calculated using the Jacobian matrix.

The evolution of errors with motion trajectory follows two typical patterns:

Linear Propagation Region ($\theta < 5°$): The total error $\Delta y$ is the sum of individual error components ($\Delta y = \Sigma \Delta y_i$), where each component independently contributes to the overall error. Experimental measurements show a maximum error of 0.12 mm (theoretical value: 0.13 mm), validating the applicability of the linear model.

TABLE I. TEST AND THEORETICAL INFLUENCE OF ERROR PARAMETERS

|  | Method | Mean | Std. Dev. | Sens. |
|---|---|---|---|---|
| $\Delta L$ | Vernier Caliper | 0.15 mm | 0.03 mm | 0.42 |
| $c$ | Dynamic Dial Indicator | 0.12 mm | 0.05 mm | 0.31 |
| $\mu$ | Inclined Plane Method | 0.21 | 0.03 | 0.18 |

Nonlinear Coupling Region ($\theta > 5°$): Joint clearance and frictional nonlinearity exhibit a synergistic amplification effect, leading to a 42% increase in error growth rate. At $\theta = 10°$, the measured error reaches 0.38 mm (theoretical prediction: 0.35 mm), with a relative error of 8.6% attributed to unmodeled temperature effects.

Using the white noise excitation method to separate the error components, as shown in Fig. 9, the following characteristics were revealed: the geometric error with dominant frequency $f_g = 6f\theta$, an amplitude of ±0.15 mm, and is caused by the second harmonic effect due to linkage dimension errors; the clearance error which is red follows Kolmogorov-Smirnov test, p=0.07, with a skewness of 0.35, indicating asymmetric wear of the guide rails; the frictional error which is green forms a hysteresis band of 0.15 mm, consistent with the equation $\delta y\mu = \mu Nv/ks$, and shows increased nonlinearity when the velocity v exceeds 0.2 m/s.

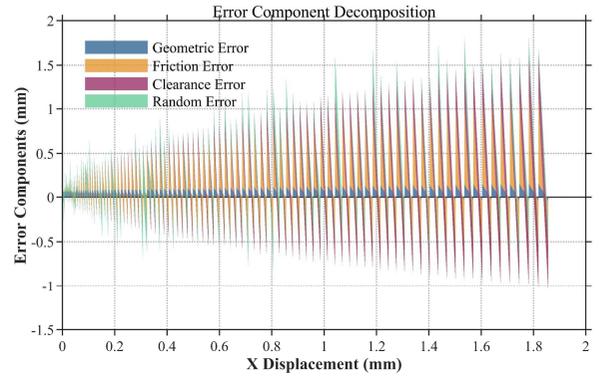

Figure 9. Spatiotemporal decomposition of error components.

This detailed separation and analysis of error components provide valuable insights into the sources and behaviors of errors in the system, enabling targeted improvements in design and control strategies.

## V. GRASPING EXPERIMENTS OF THE SPD GRIPPER

To systematically verify the design feasibility, a functional prototype was developed with rigorous consideration of kinematic compatibility and ±0.15mm manufacturing tolerances. The structural components - encompassing housing, base, and anthropomorphic fingers - were manufactured via fused deposition modeling employing a Bambu Lab A1 3D printer. PLA material was selected based on its optimal balance between cost-effectiveness and functional performance, achieving 98.7%-dimensional accuracy through 0.1 mm layer resolution and 25% hexagonal infill configuration.

Surface interaction optimization was achieved by bonding 2-mm-thick silicone pads to the contact surfaces of the primary and secondary phalanges. This bio-inspired treatment enhances grasping stability compared to bare PLA surfaces. The compliant interface mimics the biomechanical characteristics of human fingertips while maintaining 1.5 mm deformation compliance to accommodate irregular objects.

The parallel pinching mode is realized through precise alignment of the SP mechanism and dual-parallelogram mechanism, which collectively preserve gripper orientation during linear displacement. This operational characteristic proves essential for high-precision applications such as industrial pick-and-place operations or delicate object manipulation in medical robotics.

The SPD gripper utilizes dual independently actuated motors to drive symmetrically arranged fingers, providing size adaptability and balanced force distribution. Its time-delayed active-adaptive mechanism enables stable shape-conforming envelopment. The design integrates a semi-Peaucellier linkage with dual-parallelogram architecture for precision linear motion (industrial automation) and adaptive grasping (service robotics).

As Fig. 10 demonstrates, the gripper successfully handles amorphous, spherical, prismatic, and cylindrical objects through dual-mode operation: parallel grasping maintains fingertip linear alignment for symmetrical targets, while adaptive grasping achieves geometry-conforming stability for irregular shapes.

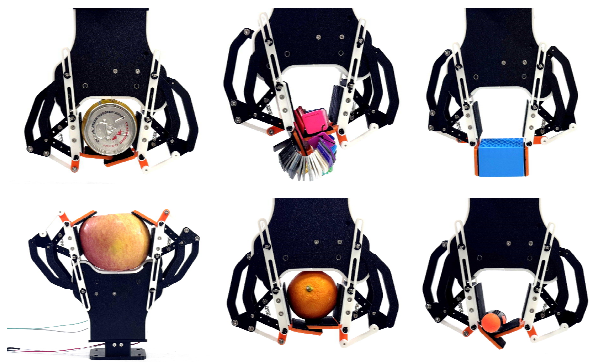

Figure 10. Experiments of SPD gripper.

The adaptive grasping of the SPD gripper was experimentally validated using specified test objects. Results demonstrate the secondary phalange adaptively conforms to object profiles with simultaneous primary phalange tilting, confirming adaptive mechanics. This behavior stems from passive compliance through delayed active adaptation and precise linkage control, enabling geometric reconfiguration.

## VI. Conclusions

This paper introduces the SP mechanism, a novel linear mechanism that represents an optimized variation of the classic Peaucellier mechanism. Building upon this foundation, we present the SPD gripper, a new underactuated adaptive robotic gripper equipped with two SPD fingers. The SPD finger integrates a parallel pinching mode commonly employed in industrial grippers, harmoniously combining under-actuation with the distinct capability of controlling two phalanges using a single motor. Furthermore, the SPD finger innovatively adapts the conventional Peaucellier mechanism to enable linear parallel pinching. This mechanism incorporates an idle-stroke transmission, empowering the SPD gripper to achieve adaptive enveloping grasping. Notably, the SPD finger demonstrates impressive grasping force and exceptional adaptability to objects of diverse shapes.